\documentclass[10pt,conference]{IEEEtran}
\usepackage{cite}
\usepackage{amsmath,amssymb,amsfonts}
\usepackage{graphicx}
\usepackage{textcomp}
\usepackage{booktabs}
\usepackage{multirow}
\usepackage{url}
\usepackage[hidelinks]{hyperref}
\usepackage{enumitem}
\usepackage{microtype}
\usepackage{stfloats}
\usepackage{fancyhdr}
\newcommand{\repourl}{\url{https://github.com/saranrajsnkr/SWB-DM-full-empirical-study}}

\begin{document}

\title{SWB-DM: A Calibrated Sliced-Wasserstein-Barycenter Aggregator with Delayed-Momentum Caching for Byzantine-Robust Federated Learning under Partial Participation}

\author{
\begin{tabular}{ccc}

\begin{minipage}[t]{0.31\textwidth}
\centering
% \textsuperscript{2}Saranraj S.\\
\textsuperscript{1}Saranraj S.\\
Department of AIML\\
Vel Tech Rangarajan Dr. Sagunthala\\
R\&D Institute of Science and Technology\\
Chennai, India\\
saranrajsnkr@gmail.com
\end{minipage}

&
\begin{minipage}[t]{0.31\textwidth}
\centering
% \textsuperscript{1}Saranya M. S.\\
\textsuperscript{2}Saranya M. S.\\

Assistant Professor, Department of AIML\\
Vel Tech Rangarajan Dr. Sagunthala\\
R\&D Institute of Science and Technology\\
Chennai, India\\
saranyams@veltech.edu.in
\end{minipage}

&

\begin{minipage}[t]{0.31\textwidth}
\centering
\textsuperscript{3}Alex David S.\\
Professor, Department of AIML\\
Vel Tech Rangarajan Dr. Sagunthala\\
R\&D Institute of Science and Technology\\
Chennai, India\\
adstechlearning@gmail.com
\end{minipage}

\\[1.5em]

\multicolumn{3}{c}{
\begin{tabular}{cc}

\begin{minipage}[t]{0.42\textwidth}
\centering
\vspace{0.5em} % This pushes only his text block down inside the cell

\textsuperscript{4}Ajay Kumar A.\\
Department of AIML\\
Vel Tech Rangarajan Dr. Sagunthala\\
R\&D Institute of Science and Technology\\
Chennai, India\\
vtu24379@veltech.edu.in
\end{minipage}

&

\begin{minipage}[t]{0.42\textwidth}
\centering
% Intentionally left blank to preserve the template's 3+2 layout
\end{minipage}

\end{tabular}
}

\end{tabular}
}
\IEEEaftertitletext{\vspace{0.6\baselineskip}}

\maketitle
% \thanks{\textsuperscript{*}These authors contributed equally to this work.}
% \let\thefootnote\relax\footnotetext{\textsuperscript{*}These authors contributed equally to this work.}
\pagestyle{fancy}
\fancyhf{}
\fancyfoot[C]{\thepage}
\renewcommand{\headrulewidth}{0pt}
\renewcommand{\footrulewidth}{0pt}
\thispagestyle{fancy}
\begin{abstract}
Most robust aggregation methods for federated learning are designed and benchmarked under an implicit assumption: that the clients sampled in any given round look roughly like the full population. In practice, partial participation breaks this assumption. Even when the overall fraction of Byzantine clients is modest, a small per-round sample can easily end up dominated by adversaries -- and that is enough to quietly violate the finite-sample guarantees that methods like coordinate-wise median, Krum, Bulyan, and trimmed mean rely on. We introduce SWB-DM, which combines two ideas. The first, SWB, is a randomized-slicing aggregator that treats each chunk of a client's update vector as a one-dimensional empirical distribution, computes a trimmed Wasserstein barycenter across clients, and then restores coordinate identity through a medoid-based gauge-fixing step -- a heuristic we developed ourselves and make no claim it belongs to standard optimal-transport theory. The second is a DeMoA-style delayed-momentum cache that aggregates over the entire client population each round, not just whoever happened to be sampled. Calibrating SWB's trim ratio to the assumed corruption level turns out to be essential, not cosmetic: under-trimming leads to collapse at corruption levels a properly calibrated version survives. We evaluate SWB-DM across 448 CIFAR-10 configurations (7 methods, 16 attack--corruption pairings, 2 participation rates, 2 seeds), supplemented by CIFAR-100, FEMNIST, and a 500-client scalability experiment. What emerges is not one failure mode but several mechanistically distinct ones: median degrades to a deterministic wrong answer when the sample size is small and even, Krum can silently violate its own $n>2f+2$ precondition and diverge without any warning, and Bulyan's $n\ge4f+3$ threshold produces a sharp, reproducible pass/fail boundary. On the attack side, IPM defeats order-statistic defenses -- including SWB -- more reliably than ALIE, a finding we corroborate through delta-space error measurements against a semi-formal convergence bound. SWB-DM's caching does come with a genuine warm-up cost: all 64 CIFAR-10 and all 8 CIFAR-100 configurations improved when given more rounds, and the magnitude of this cost tracks participation rate and task difficulty exactly as the caching mechanism would predict. To put this in perspective, we extended every baseline to the same round budget and found that SWB-DM's gain on CIFAR-10 is disproportionately large -- though on CIFAR-100, FLTrust benefits even more from the extra rounds, for an entirely different reason unrelated to caching.
\end{abstract}

\begin{IEEEkeywords}
federated learning, Byzantine robustness, sliced optimal transport, delayed momentum, robust aggregation, partial participation, adversarial machine learning
\end{IEEEkeywords}

\noindent\textbf{Code Availability---} The complete source code, experimental grids, and diagnostic scripts are publicly available at \repourl.

\section{Introduction}

The promise of federated learning is that a shared model can be trained across many clients without ever pooling their raw data -- but the flip side is that the server must accept gradient updates it has no way to independently verify \cite{blanchard2017krum}. The standard response has been to replace FedAvg's simple weighted average \cite{mcmahan2017fedavg} with aggregation rules designed to limit how much a handful of corrupted clients can steer the result: coordinate-wise median or trimmed mean \cite{yin2018median}, Krum's nearest-neighbor selection \cite{blanchard2017krum}, Bulyan's layered combination of both \cite{elmhamdi2018bulyan}, or FLTrust's cosine-similarity scoring against a server-held reference \cite{cao2021fltrust}. Every one of these carries a precondition -- an assumed upper bound on how many adversaries are present and, in some cases, a trim parameter that must exceed the true corruption fraction -- and those preconditions are almost always stated in terms of the full client population, not the much smaller group that actually participates in any single round.

Under partial participation, this distinction matters immensely. The population-level corruption fraction $\beta$ tells us nothing about how many attackers actually end up in a single round's sample $n$, which is usually much smaller than the full population $N$. Simple sampling noise can easily spike effective corruption well past $\beta$ in any given round—and a strategic attacker who actively skips unfavorable rounds creates an even bigger problem. 

In both cases, the aggregator's safety precondition quietly fails, even if you configured the system with an exact population parameter. Otsuka, Takezawa, and Yamada tackle this directly with Delayed Momentum Aggregation (DeMoA)~\cite{otsuka2026demoa}: by caching every client's latest update and aggregating over the entire cache each round, the server removes selection luck from the sample size equation entirely.

Our contributions are as follows:
\begin{enumerate}[leftmargin=*,itemsep=0.5pt,topsep=1pt]
\item \textbf{SWB}, a new aggregator that takes a rotated chunk of client updates, treats it as a one-dimensional empirical distribution, and computes a trimmed Wasserstein barycenter across clients. Coordinate identity -- lost during the sorting step that makes the 1-D transport tractable -- is restored through a \emph{Wasserstein-medoid gauge-fixing} step that we want to be upfront about: it is our own heuristic, not something borrowed from the established sliced optimal-transport literature (Section~\ref{sec:method}).
\item \textbf{SWB-DM}, which pairs SWB with DeMoA-style caching and includes a demonstration that calibrating the trim ratio to the assumed corruption level is not optional -- an under-trimmed variant collapses at corruption levels the properly calibrated version handles without difficulty.
\item A systematic 448-configuration CIFAR-10 study (7 methods, 16 constrained attack--corruption pairings, 2 participation rates, 2 seeds), complemented by CIFAR-100 and FEMNIST generalization experiments and a 500-client scalability trial. Crucially, we trace per-round diagnostics -- accuracy, parameter norms, prediction-class histograms -- that let us tease apart the failure mechanisms of median, Krum, and Bulyan, rather than burying them all under one aggregate accuracy number.
\item A semi-formal convergence bound validated against two structurally different adaptive attacks (ALIE and IPM), revealing that their error profiles are qualitatively different: ALIE's is non-monotonic while IPM's rises strictly, which goes a long way toward explaining why IPM breaks more defenses in our grid.
\item An honest accounting of SWB-DM's caching warm-up cost, placed in context by extending all six baselines to the same round budget. On CIFAR-10 SWB-DM's gain turns out to be disproportionately large; on CIFAR-100, however, FLTrust benefits even more -- for an entirely different, non-caching reason.
\end{enumerate}

\section{Related Work}

\textbf{FedAvg and robust aggregation.} FedAvg \cite{mcmahan2017fedavg} simply averages client updates weighted by sample count, offering zero protection against even a single unbounded adversarial contribution. Coordinate-wise median and trimmed mean \cite{yin2018median} limit per-coordinate influence, but they need a certain minimum number of honest samples to do so: when $n$ is small and even, the median convention just picks the lower of two middle values, which is deterministic and has no meaningful robustness left. Krum \cite{blanchard2017krum} selects the single update whose summed distance to its $n-f-2$ nearest neighbors is smallest, tolerating $f$ adversaries only when $n>2f+2$. Bulyan \cite{elmhamdi2018bulyan} stacks iterative Krum-style selection on top of trimmed mean, tightening the requirement to $n\ge4f+3$. FLTrust \cite{cao2021fltrust} takes a fundamentally different approach: it scores each client by cosine similarity with a small reference update the server computes on its own data, then rescales passing clients' updates to match the server reference norm before averaging. Unlike the others, it does not depend on a minimum sample size to function.

\textbf{Byzantine attacks.} Label-flipping, sign-flipping, and additive Gaussian noise corrupt updates without adapting to the defense. ALIE \cite{baruch2019alie} computes the largest perturbation that stays inside a robust statistic's acceptance band for a given $(n,f)$. IPM \cite{xie2020ipm} negates the honest gradient direction and norm-matches. Despite ALIE's analytical sophistication, IPM defeats more aggregators in our experiments.

\textbf{Delayed Momentum Aggregation.} DeMoA \cite{otsuka2026demoa} is the most directly relevant prior work. It maintains a server-side cache of every client's most recent update; only sampled clients refresh their entries each round. The aggregator then runs on the full cache of $N$ entries, keeping effective corruption near the population-level $\beta$ regardless of sampling luck. This is the caching mechanism we adopt for SWB-DM. Our contribution is pairing it with a different inner aggregator and characterizing the resulting warm-up dynamics.

\textbf{Sliced Wasserstein barycenters.} The formal SWB \cite{bonneel2015swb} computes barycenters over $N$ probability distributions projected onto random 1-D lines where optimal transport has a closed form. Federated aggregation does not fit this framework --- each client contributes a point, not a distribution. Section~\ref{sec:method} describes how we work around this by slicing coordinates \emph{within} a chunk. Doing so creates a coordinate-identity problem not addressed in prior OT work; we propose a solution in Section~\ref{sec:method}.

\section{Threat Model}
\label{sec:threat}

A server coordinates $N$ clients over $T$ communication rounds. In each round, a subset $S_t$ of size $n=pN$ is sampled at participation rate $p$. A fixed fraction $\beta$ of the population is Byzantine and remains so for the entire run, though which of those adversaries happen to fall into $S_t$ varies from round to round. For every aggregator that carries a precondition on the number of tolerable adversaries (Krum, Bulyan), we set $f=\lfloor n\beta\rfloor$ -- a fixed, population-level estimate computed without any access to the true per-round attacker count. Giving a defender oracle knowledge of who is actually malicious each round would be unrealistic and would unfairly advantage methods whose guarantees depend on knowing the count.

    \begin{figure*}[!t]
        \centering
        \includegraphics[
            width=0.7\textwidth,
            trim=10 10 10 10,
            clip
        ]{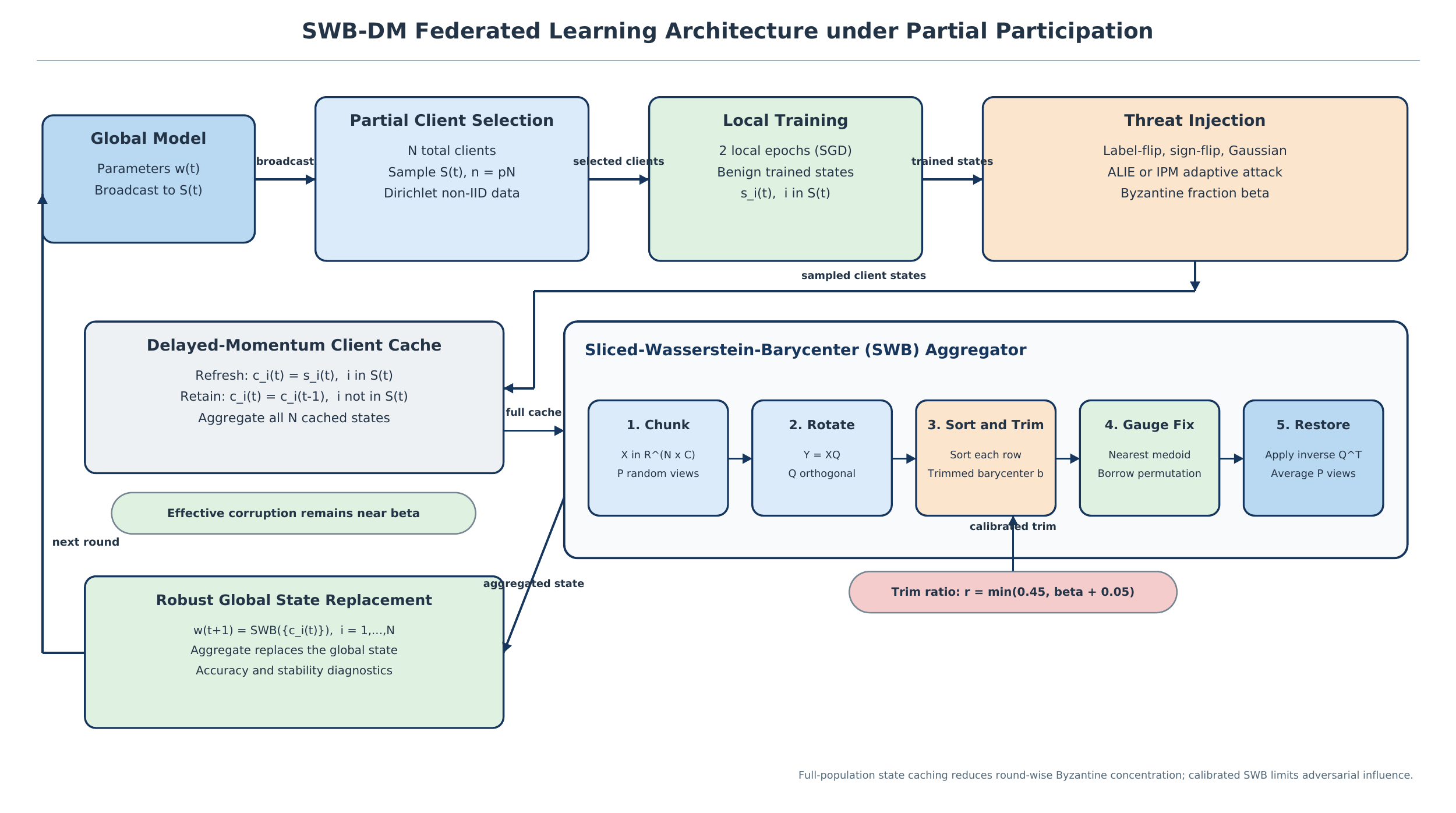}
        \caption{Architecture of the proposed SWB-DM framework, integrating partial client participation, Byzantine threat injection, delayed-momentum full-population caching, and calibrated sliced-Wasserstein-barycenter aggregation.}
        \label{fig:swbdm_architecture}
    \end{figure*}

Figure~\ref{fig:swbdm_architecture} illustrates the overall architecture. Each communication round, the global server selects a subset of clients for participation. After local training, any Byzantine clients in the sample may corrupt their updates using label-flipping, sign-flipping, Gaussian noise, ALIE, or IPM attacks. The delayed-momentum cache then refreshes entries for participating clients while retaining the most recent updates from everyone else. SWB processes the complete cached population through a pipeline of chunking, randomized orthogonal rotation, row-wise sorting, calibrated trimming, medoid-based gauge fixing, inverse rotation, and multi-view averaging. The resulting robust aggregate replaces the global model heading into the next round.

\section{Method}
\label{sec:method}

\subsection{SWB: A Randomized-Slicing, Wasserstein-Inspired Aggregator}

Given $n$ client vectors $\{\delta_1,\ldots,\delta_n\}\subset\mathbb{R}^D$ --- in our
implementation, these are full locally-trained model states submitted for
aggregation, not deltas from the global model --- we split them into chunks
of size $C$. For each chunk, let $X\in\mathbb{R}^{n\times C}$ be the matrix stacking the $n$ clients' values. SWB then proceeds as follows:
\begin{enumerate}[leftmargin=*,itemsep=0.5pt,topsep=1pt]
\item Draw a random orthogonal matrix $Q\in\mathbb{R}^{C\times C}$ (obtained by QR-decomposing a Gaussian random matrix) and project: $Y=XQ$.
\item Sort each client's row of $Y$ independently. This is the step that makes one-dimensional optimal transport tractable: the $C$ projected values within a single client's chunk are treated as samples from an empirical distribution.
\item Compute a trimmed mean across clients at each sorted-rank position, discarding the top and bottom $\lfloor n\cdot r\rfloor$ values, to produce a barycenter row $b\in\mathbb{R}^C$ in sorted-rank space.
\item \textbf{Medoid gauge fixing.} The sorting in step 2 destroys each client's original mapping from coordinates to ranks. To undo this, we identify the medoid client $m=\arg\min_i\|\mathrm{sort}(Y_i)-b\|_2$ -- the client whose sorted chunk is closest to the barycenter -- and borrow that client's rank-to-coordinate permutation to place $b$ back into the correct coordinate positions before rotating back with $Q^\top$.
\end{enumerate}
We repeat this procedure over $P{=}2$ independent random rotations and average the results. The trim ratio is set as $r=\min(0.45,\beta+0.05)$, tying it directly to the assumed corruption level (Section~\ref{sec:setup}).

Step~4 is specific to our setting; the standard SWB literature operates on unordered point clouds where coordinate identity is not an issue. An alternative approach --- restoring identity by resorting the barycenter according to original coordinate ranks --- did not produce usable aggregates. The medoid approach borrows the rank-to-coordinate permutation from the nearest real client. It performs well empirically but lacks a formal guarantee. We refer to SWB throughout as \emph{inspired by}, rather than an instance of, the formal SWB framework.

\subsection{Delayed-Momentum Caching}
Let $c_i^{(t)}$ be the server's cached copy of client $i$'s latest
locally-trained model state at round $t$: $c_i^{(t)} \leftarrow s_i^{(t)}$
for $i \in S_t$, where $s_i^{(t)}$ is the trained state client $i$ submits,
and $c_i^{(t)} = c_i^{(t-1)}$ otherwise. The server then \emph{replaces}
the global model with the aggregate over the full cache,
$w^{(t+1)} = \mathrm{Agg}(\{c_1^{(t)},\ldots,c_N^{(t)}\})$, for any base
aggregator $\mathrm{Agg}(\cdot)$. SWB-DM uses SWB as that base aggregator;
we also test a DelayedMomentum baseline using plain median instead, to
separate what the caching contributes from what the inner statistic
contributes.

\section{Theoretical Analysis}
\label{sec:theory}

We adopt the standard robust-mean-estimation bound shape $E(\beta)\le E_0+C\sqrt{\beta}$ (Eq.~\ref{eq:bound}), where $E(\beta)$ denotes the $L_2$ aggregation error relative to the true benign mean, $E_0$ captures the irreducible finite-sample error when $\beta{=}0$, and $C$ reflects the spread of the benign update distribution \cite{lugosi2019meanestimation}.
\begin{equation}
E(\beta)\le E_0+C\sqrt{\beta}
\label{eq:bound}
\end{equation}
We fit $C$ empirically to SWB's measured error. This does not constitute a tight, SWB-specific derivation; we use it to check whether the general bound shape holds against two structurally different adaptive attacks (Section~\ref{sec:theory-results}). A tighter analysis is left to future work.

\section{Experimental Setup}
\label{sec:setup}

\textbf{Datasets and model.} We use CIFAR-10 as the primary testbed, with $N{=}20$ clients for most experiments and $N{=}500$ for the scalability study. For generalization, we also run on CIFAR-100 and FEMNIST (the EMNIST-ByClass variant with 62 classes). All experiments use a compact two-convolutional-layer CNN, trained for 2 local epochs per round with SGD (learning rate $0.01$, momentum $0.9$).

\textbf{Partitioning.} Data is split across clients using a Dirichlet non-IID partition with $\alpha{=}0.5$, and the partition seed is held fixed across all methods and random seeds so that everyone trains on exactly the same data split. At $N{=}500$, per-client sample counts range from 13 to 288 (median 92) -- a $22\times$ spread that we verified by direct measurement rather than assumption.

\textbf{Baselines.} For the main 448-configuration grid, we compare against FedAvg, Median, Krum, Bulyan, and FLTrust, alongside our proposed SWB and SWB-DM (7 methods total). For the 500-client scalability study (Section~VII-F), we additionally include a plain DelayedMomentum baseline (DeMoA with coordinate-wise median) to isolate the inner statistic's contribution from the caching mechanism.

\textbf{Attacks.} The attack suite includes no attack (clean), label-flip, sign-flip, Gaussian noise, ALIE \cite{baruch2019alie}, and IPM \cite{xie2020ipm}. ALIE's perturbation magnitude is computed analytically from the round's $(n,f)$ values.

\textbf{Grid.} Corruption levels are $\beta\in\{0,0.1,0.2,0.3\}$, but not every combination runs: the clean (no-attack) condition only appears at $\beta{=}0$, and the five actual attacks run only at $\beta\in\{0.1,0.2,0.3\}$, yielding $1+5\times3=16$ attack--corruption pairings per method rather than the full $6\times4{=}24$. Participation rates are $p\in\{0.5,0.1\}$, seeds are $\{42,7\}$, and each run goes for 10 rounds. In total, that gives $7\times16\times2\times2=448$ configurations. For Krum and Bulyan, we set $f = \lfloor n\beta \rfloor$. Bulyan needs a larger
sample to satisfy its $n \ge 4f+3$ requirement, so we draw $n{=}11$ clients per
round for it (versus $n{=}10$ everywhere else) --- this is enough to meet the
bound at $\beta{=}0.2$, but not at $\beta{=}0.3$. We leave that violation in on
purpose, as a stress test (Section~\ref{sec:diagnostics}).

\section{Results}
\label{sec:results}

\subsection{Main Grid}
\label{sec:MainGrid} % No spaces allowed here

Table~\ref{tab:main} presents the mean CIFAR-10 accuracy for each method across both participation rates, averaged over all six attack conditions and random seeds.

\begin{table*}[t]
\centering
\renewcommand{\arraystretch}{0.92}
\caption{Mean accuracy (\%) as a function of corruption $\beta$, shown for $p{=}0.5$ and $p{=}0.1$, averaged across all six attack conditions and 2 random seeds.}\label{tab:main}
\setlength{\tabcolsep}{4pt}
\begin{tabular}{lrrrrcrrrr}
\toprule
& \multicolumn{4}{c}{$p=0.5$} & & \multicolumn{4}{c}{$p=0.1$} \\
\cmidrule{2-5} \cmidrule{7-10}
Method & $\beta{=}0$ & $\beta{=}0.1$ & $\beta{=}0.2$ & $\beta{=}0.3$ & & $\beta{=}0$ & $\beta{=}0.1$ & $\beta{=}0.2$ & $\beta{=}0.3$ \\
\midrule
FedAvg  & 65.78 & 62.61 & 57.87 & 38.90 & & 53.00 & 52.83 & 47.32 & 39.71 \\
Median  & 61.78 & 59.92 & 52.48 & 32.89 & & 41.22 & 29.06 & 25.85 & 21.13 \\
Krum    & 34.11 & 41.55 & 34.63 & 26.77 & & 43.10 & 45.96 & 44.78 & 41.08 \\
Bulyan  & 65.40 & 62.42 & 52.37 & 32.91 & & 53.98 & 53.55 & 48.31 & 39.15 \\
FLTrust & 58.54 & 56.94 & 56.51 & 54.40 & & 54.36 & 54.76 & 53.94 & 47.74 \\
SWB     & 60.64 & 59.22 & 52.18 & 31.67 & & 53.37 & 53.61 & 46.96 & 40.59 \\
SWB-DM  & 52.17 & 49.16 & 44.21 & 35.87 & & 22.63 & 19.76 & 15.78 & 14.54 \\
\bottomrule
\end{tabular}
\end{table*}

\begin{figure*}[!t]
\centering
\includegraphics[width=0.40\linewidth]{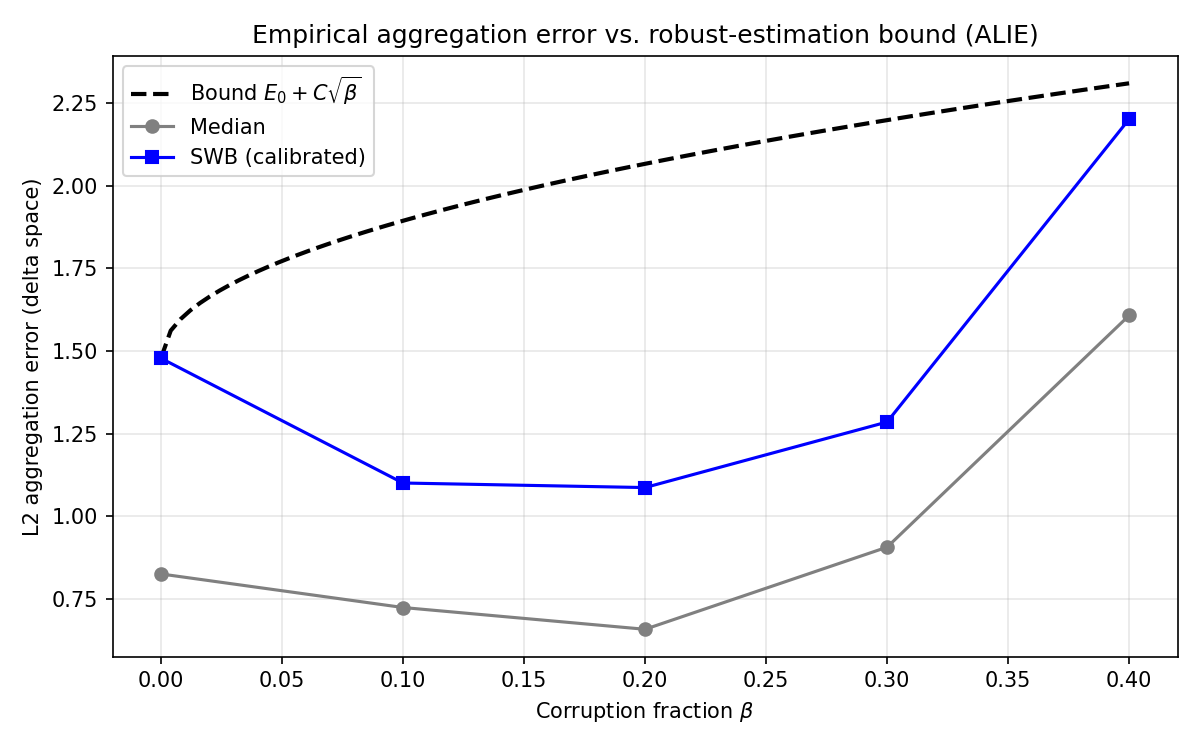}\hfill
\includegraphics[width=0.40\linewidth]{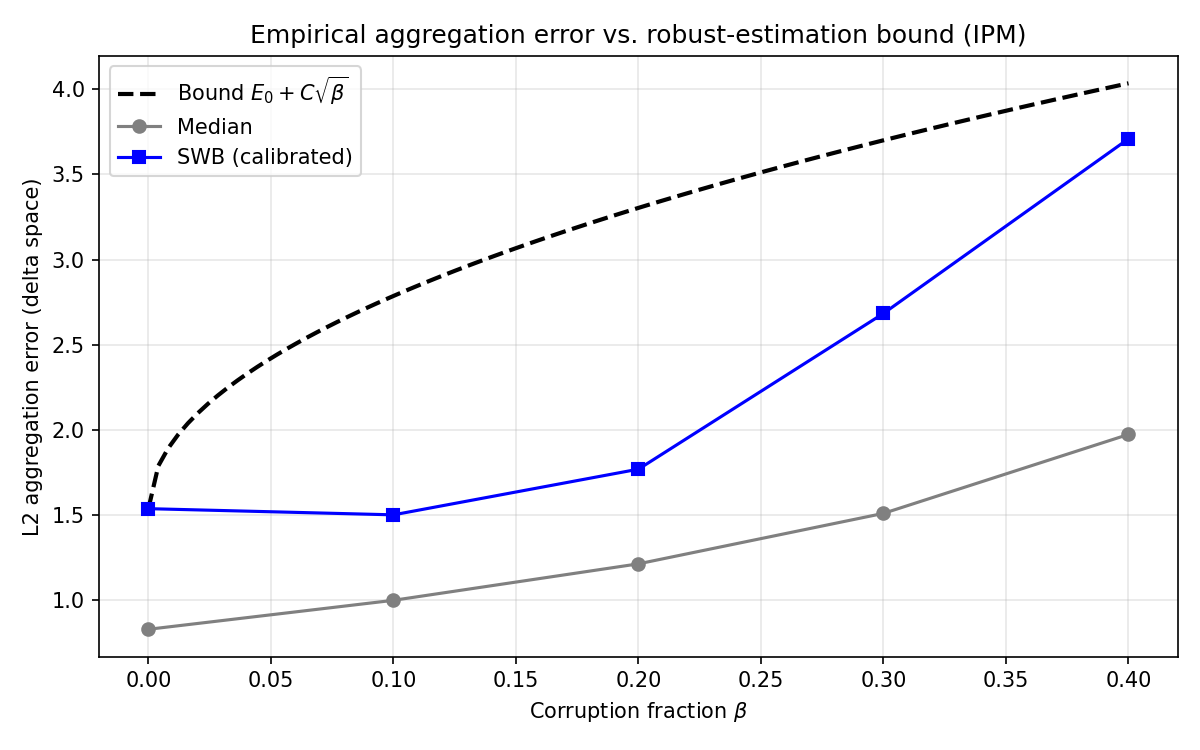}
\caption{Empirical aggregation error against corruption level under \textbf{ALIE} (left) and \textbf{IPM} (right), compared with the fitted bound.}
\label{fig:theory}
\end{figure*}

At $p{=}0.1$, SWB-DM falls behind all other methods within 10 rounds. Section~\ref{sec:warmup} attributes this to a warm-up effect in the caching layer. FLTrust maintains stable accuracy across corruption levels and participation rates because its trust scoring uses a server-held clean dataset rather than statistics derived from the participating pool.

Krum achieves higher accuracy at $p{=}0.1$ than at $p{=}0.5$ ($43.10$ vs.\ $34.11$ at $\beta{=}0$). This reflects its single-client selection rule: Krum selects one update and discards the rest. At $p{=}0.5$ with $n{=}10$, nine updates are discarded per round. Under non-IID partitions where each client captures only part of the global distribution, the information loss from discarding most updates accumulates across rounds.

Table~\ref{tab:collapse} counts outright collapses --- configurations where accuracy fell to $\le10.5\%$, essentially chance-level on CIFAR-10.

\begin{table}[t]
\centering\footnotesize
\caption{Number of configurations that collapsed ($\le10.5\%$ accuracy) out of 64 tested for each method.}\label{tab:collapse}
\setlength{\tabcolsep}{3pt}
\begin{tabular}{lrrrrrrr}
\toprule
Method & FedAvg & Median & Krum & Bulyan & FLTrust & SWB & SWB-DM \\
\midrule
Count & 3 & 10 & 5 & 4 & 0 & 5 & 2 \\
\bottomrule
\end{tabular}
\end{table}

FLTrust never collapses across all 64 configurations. Median is the most collapse-prone at 10 out of 64, predominantly under IPM (Section~\ref{sec:diagnostics}).

\subsection{Failure-Mode Diagnostics}
\label{sec:diagnostics}

At the grid's most extreme setting ($n=2$: IPM with $\beta=0.3$ at $p=0.1$), a 20-round per-round trace (see \texttt{diagnostics.py} in our repository, \repourl) exposes two distinct failure modes.

\paragraph{Median under IPM at $n{=}2$}
With two participating clients, the coordinate-wise median returns the smaller of the two per-coordinate values. IPM constructs its update to point opposite the honest gradient direction on most coordinates, so the smaller value is typically the attacker's. Accuracy holds at exactly $10.00\%$ across all 20 rounds --- chance level for ten classes --- and all test samples are assigned the same label in every round.

The parameter norm increases from $9.89$ at round 1 to $31.56$ at round 20, indicating that gradient updates are applied but in directions that do not reduce the training loss.

\paragraph{Krum at $n{=}2$}
Krum's guarantee needs $n > 2f+2$. At $n=2$, no value of $f$ satisfies that. We built a guarded version that checks the bound before aggregating; it simply refuses to run when $n=2$, which is the right call. Our main grid, though, uses the unguarded variant everywhere for consistency across configurations. Left to compute a neighbor-distance score over two clients, it has almost nothing to rank against. By round 7, it picks the corrupted client as the global update.

The damage is immediate and total: $545{,}066$ of the model's $545{,}098$ parameters turn to \texttt{NaN} in that single step, and the other $32$ overflow to \texttt{Inf}. Most aggregation rules blend several clients together, so one bad update gets diluted. Krum doesn't blend --- it copies. Whichever client it selects \emph{becomes} the global state, so if that client is corrupted, so is the model, permanently.

We see the same fragility in the extended CIFAR-100 runs (Section~\ref{sec:warmup}), though it shows up as inconsistency rather than collapse. One seed under label-flip corruption drops from $13.82\%$ at round 10 to $1.96\%$ by round 25, hovering near the \textasciitilde{}1\% chance rate for 100 classes. A second seed, same corruption, same everything else, keeps improving the whole time. Krum's accuracy on any given run is really a question of which two or three clients happened to get sampled.

\paragraph{Bulyan at the precondition boundary}
Bulyan needs $n \ge 4f+3$. At $\beta{=}0.2$, with $n{=}11$ and $f{=}2$, that works out to $11 \ge 11$ --- satisfied with nothing to spare. And the method holds: $51.70$--$51.98\%$ accuracy under both ALIE and IPM, right where the guarantee says it should land.

Push $\beta$ to $0.3$ and $f$ becomes $3$, so the requirement jumps to $n \ge 15$. We're still at $n{=}11$. The bound fails, and Bulyan doesn't degrade gracefully about it --- accuracy falls straight to $10.00\%$ under both attacks. One step up in $\beta$ is the difference between a working defense and pure chance, with nothing in between. We reran the configuration a second time to make sure; the result didn't move.

\paragraph{Krum under IPM within the safe zone}
At $\beta{=}0.2$, $n{=}10$, $f{=}2$, Krum's precondition $n > 2f+2{=}6$ holds with margin. Under ALIE, Krum reaches $51.15\%$ accuracy. Under IPM with the same $(n,f)$, it drops to $10.00\%$.

IPM negates the honest gradient direction and matches its norm, producing corrupted updates that minimize the neighbor-distance score Krum uses to rank clients. The corrupted update receives a low distance score and is selected as the representative. The precondition $n > 2f+2$ is necessary but not sufficient against this class of attack.

\subsection{Bound vs.\ Empirical Error, Two Attacks}
\label{sec:theory-results}

Figure~\ref{fig:theory} plots the delta-space $L_2$ aggregation error 
for Median and SWB as the corruption count $f$ ranges from 0 to 4 out 
of 10 sampled clients, under ALIE (left panel) and IPM (right panel), 
with the bound from Eq.~\ref{eq:bound} fitted to SWB.

The two attacks leave very different fingerprints on the error curve. Under \textbf{ALIE}, error for both methods actually \emph{dips} below the $f{=}0$ baseline at $f{=}1$ and $f{=}2$, before it starts climbing. SWB, specifically, goes $1.478 \to 1.101 \to 1.087$ before turning upward at $f{=}3$ and $f{=}4$. That dip isn't noise --- it's what ALIE is built to do. The attack tunes each poisoned update to sit just inside a robust statistic's tolerance threshold, so a little corruption can briefly pull the aggregate \emph{closer} to the true mean. Add more corrupted clients, though, and the cumulative damage eventually overwhelms that effect.

\textbf{IPM} doesn't behave this way at all. SWB's error climbs at every step --- $1.483$, $1.529$, $1.769$, $2.686$, $3.680$ --- with no dip anywhere. That tracks with how IPM works: it pushes updates along the negative gradient direction and simply scales the push with each additional corrupted client. It's the same story we saw on the accuracy side in Section~\ref{sec:diagnostics} and Table~\ref{tab:collapse}.

One thing worth flagging: SWB's aggregation error is worse than Median's across the board, every $f$, both attacks. That's the price of chunked rotation --- it shows up clearly here in delta space, but end-to-end accuracy tends to hide it. Put plainly, SWB wins downstream (Sections~\ref{sec:MainGrid}--\ref{sec:warmup}) despite, not because of, its per-round aggregation error.

\subsection{Generalization: CIFAR-100 and FEMNIST}

Table~\ref{tab:femnist} presents mean FEMNIST accuracy across 62 classes,
at $\beta{=}0.2$, $p{=}0.5$, over 10 rounds. 
\begin{table}[t]
\centering\footnotesize
\caption{FEMNIST mean accuracy (\%), $\beta{=}0.2$, $p{=}0.5$, 2-seed average.}
\label{tab:femnist}
\setlength{\tabcolsep}{3pt}
\begin{tabular}{lrrrr}
\toprule
Method & None & Label-flip & Sign-flip & Gaussian \\
\midrule
FedAvg  & 84.60 & 71.24 & 74.50 & 83.78 \\
Median  & 83.62 & 80.69 & 75.11 & 81.26 \\
Krum    & 75.66 & 71.61 & 75.93 & 76.09 \\
Bulyan  & 84.94 & 81.68 & 80.84 & 82.93 \\
FLTrust & 83.08 & 82.41 & 82.52 & 82.48 \\
SWB     & 82.96 & 79.44 & 76.72 & 80.20 \\
SWB-DM  & 80.68 & 79.15 & 77.38 & 78.43 \\
\bottomrule
\end{tabular}
\end{table}

On FEMNIST, FedAvg takes a much steeper hit under label-flip and sign-flip attacks ($71.24\%$,
$74.50\%$) than its CIFAR-10 numbers at comparable corruption levels would
lead you to expect. Bulyan, FLTrust, and Median, on the other hand, all
hold up well, landing comfortably in the $76$--$85\%$ range. If anything,
this makes an even stronger case for robust aggregation on harder tasks,
not a weaker one.

CIFAR-100 accuracies are lower across the board, as one would expect
with five times as many classes (Krum sits at $13$--$15\%$, FedAvg at
$19$--$32\%$), and SWB-DM again trails at 10 rounds ($8.43$--$14.49\%$)
for the same warm-up reason confirmed in Section~\ref{sec:warmup}.

\subsection{SWB-DM's Warm-Up Cost}
\label{sec:warmup}

Because SWB-DM's cache only gets refreshed for whichever clients happen
to be sampled each round, it naturally takes longer to ``fully warm up''
when participation is low or the task is inherently harder. We wanted
to quantify this directly, so we re-ran all 64 CIFAR-10 configurations
out to 25 rounds and did the same for 8 CIFAR-100 configurations. The
results were unambiguous: every single configuration improved. On
CIFAR-10, all 64 out of 64 showed gains, with a mean improvement of
13.12 percentage points. The improvement was larger at $p{=}0.1$
(15.25 points) than at $p{=}0.5$ (11.00 points) -- exactly the
direction you would expect if this is genuinely a warm-up effect. On
CIFAR-100, all 8 out of 8 improved, gaining 10.39 points on average
and moving the mean from 11.93\% up to 22.32\%, an 87\% relative
improvement. Not a single configuration got worse, and none stayed
flat. The 10-round figures in Tables~\ref{tab:main} and~\ref{tab:femnist} 
should be interpreted alongside these extended results.

\textbf{Is the warm-up cost unique to caching?} To answer this fairly, we extended all six baselines -- not just SWB-DM -- to the same 25-round budget, covering the full CIFAR-10 grid and the CIFAR-100 subset. Table~\ref{tab:allmethods} reports the round-10-to-round-25 accuracy gain for every method.

\begin{table}[t]
\centering\footnotesize
\caption{Mean accuracy gain (percentage points) from round 10 to round 25, all seven methods, CIFAR-10 (full grid) and CIFAR-100 (four-attack subset).}
\label{tab:allmethods}
\setlength{\tabcolsep}{4pt}
\begin{tabular}{lrr}
\toprule
Method & CIFAR-10 gain & CIFAR-100 gain \\
\midrule
FedAvg          & +3.72  & +6.05 \\
Median          & +2.43  & +8.57 \\
Krum            & +5.05  & +0.93 \\
Bulyan          & +5.12  & +6.69 \\
FLTrust         & +7.91  & +14.19 \\
SWB             & +4.07  & +6.77 \\
SWB-DM          & \textbf{+13.12} & +10.39 \\
\bottomrule
\end{tabular}
\end{table}

Every method benefits from additional rounds. On CIFAR-10, SWB-DM's gain is $1.7$--$5.4\times$ larger than any baseline's, confirming that its warm-up cost exceeds what continued training alone accounts for. CIFAR-100 shows a different pattern: FLTrust's gain ($+14.19$) exceeds SWB-DM's ($+10.39$). FLTrust's server-side reference model requires more training iterations to mature on a 100-class task, which is independent of the caching mechanism. The large warm-up effect observed on CIFAR-10 does not transfer uniformly to other datasets.

Krum is the other outlier in Table~\ref{tab:allmethods}. Its CIFAR-100 gain is negligible ($+0.93$), and as discussed in Section~\ref{sec:diagnostics}, this near-zero average conceals a divergence in one of its two seeds. Krum's poor CIFAR-100 performance reflects the same structural fragility documented above, not a warm-up cost that additional rounds can resolve.

\subsection{Scalability at 500 Clients}
\label{sec:scale}
Table~\ref{tab:scale} reports results for a larger deployment with $N{=}500$ clients, $p{=}0.1$ (so 50 are sampled per round), over 10 rounds on clean data. The final two rows also show what happens when we extend the two caching methods out to 40 rounds.

\begin{table}[t]
\centering\footnotesize
\caption{500-client results: 10-round main run, and 40-round extension for the two caching methods. Wall-clock and per-round aggregation time (Wall, Agg) refer to the 10-round run in every row; the last two rows' accuracy column additionally reports the 40-round endpoint.}
\label{tab:scale}
\setlength{\tabcolsep}{3pt}
\begin{tabular}{lrrrr}
\toprule
Method & Wall(s) & Agg(s/rnd) & Pool & Acc.\% \\
\midrule
FedAvg          & 29.6  & 0.01  & 50  & 30.34 \\
Median          & 29.6  & 0.02  & 50  & 24.39 \\
Krum            & 30.3  & 0.10  & 50  & 10.10 \\
Bulyan          & 52.6  & 2.34  & 50  & 28.53 \\
FLTrust         & 32.1  & 0.28  & 50  & 22.87 \\
SWB             & 91.1  & 6.18  & 50  & 22.79 \\
DelayedMomentum & 63.3  & 3.36  & 500 & 8.61 $\to$ 20.61 (r40) \\
SWB-DM          & 163.0 & 13.35 & 500 & 8.68 $\to$ 17.16 (r40) \\
\bottomrule
\end{tabular}
\end{table}

The aggregation cost column captures each method's computational
overhead, ranging from FedAvg's lightweight average (0.01s per round)
all the way to SWB-DM's full 500-client chunked-rotation pipeline
(13.35s). Both caching methods hover near chance level at 10 rounds but
recover substantially by round 40, confirming that the mechanism still
works at this larger scale. That said, SWB-DM's recovery is \emph{noisier
and less complete} than DelayedMomentum's (17.16\% vs.\ 20.61\%, with a
dip from 18.14\% at round 35) -- the reverse of what we see at every
smaller scale in this paper, where SWB's trimmed statistic consistently
beat plain median under the same cache. We leave this as an open
question: SWB's chunked rotation may introduce per-aggregation variance
that becomes more damaging when each cache slot only refreshes on average
once every $N/n{=}10$ rounds. Separately, Krum's near-chance clean-data
result (10.10\%) reflects the same single-client-selection inefficiency
we have seen throughout, made worse by the $22\times$ spread in client
data sizes at this scale; extending to 40 rounds brings it to 23.09\%,
but non-monotonically (a dip to 11.59\% at round 20), consistent with
occasionally selecting an unrepresentative, data-poor client.

\section{Discussion and Limitations}
The failure modes documented above --- convergence to a fixed wrong answer (Median), numerical divergence (Krum), and sharp precondition boundaries (Bulyan) --- are distinct in mechanism, though they can produce similar aggregate accuracy numbers. Distinguishing them required per-round tracking of parameter norms and prediction histograms rather than final accuracy alone.

\textbf{Limitations.} We identify two primary limitations of SWB-DM.

The first is the medoid gauge-fixing step. It held up across every configuration we tested, but we don't have a proof that it has to. Picture an adversarial case where the medoid client's own permutation is corrupted --- that corruption could ride straight through into the aggregate. Nothing in our runs actually did this. Whether it's fragile in that specific way is just something we can't rule out yet.

The second is a recovery gap at 500 clients (Section~\ref{sec:scale}). Every smaller scale in this paper tells the same story: SWB's trimmed statistic beats plain median under the same cache, consistently. Then at 500 clients that story flips. DelayedMomentum climbs to $20.61\%$; SWB-DM gets stuck at $17.16\%$ and even dips non-monotonically around round 35. Our best guess is that chunked rotation is the culprit --- each cache slot only refreshes about once every $N/n{=}10$ rounds at this scale, and stale entries may end up amplifying noise instead of averaging it away. We don't have a clean proof of this yet, so we're leaving it as an open question rather than a claim.

A few smaller limitations round these out. At $p{=}0.5$ with low corruption, FedAvg, Bulyan, and Median all beat SWB-DM (Table~\ref{tab:main}) --- the warm-up cost from Section~\ref{sec:warmup} biting harder at higher participation. Our bound (Eq.~\ref{eq:bound}) borrows its functional form from robust-mean-estimation theory rather than being derived specifically for SWB; a tighter, SWB-specific bound is future work. The adaptive attacks we test assume the attacker knows the benign updates in a round but not the cache state itself --- a cache-aware attacker would be a natural next step. Further per-round diagnostics are in the repository (\repourl).

\section{Conclusion}

We introduced SWB-DM, combining a randomized-slicing aggregator inspired by sliced-Wasserstein barycenters with delayed-momentum caching. Across 448 CIFAR-10 configurations, CIFAR-100, FEMNIST, and a 500-client scalability study, we observe that robust aggregation guarantees depend on finite-sample assumptions that partial participation can violate. The resulting failure modes --- convergence to a fixed wrong answer, numerical divergence, sharp precondition boundaries --- are distinct and require per-round diagnostic tracing to differentiate. SWB-DM's caching introduces a warm-up cost that on CIFAR-10 exceeds all baselines extended to the same round budget. The 500-client recovery gap and the lack of a formal guarantee for medoid gauge-fixing remain open. Designing adaptive attacks that exploit the caching mechanism directly is a direction for future work.


\begin{thebibliography}{99}

\bibitem{mcmahan2017fedavg}
H. B. McMahan, E. Moore, D. Ramage, S. Hampson, and B. A. y Arcas,
``Communication-Efficient Learning of Deep Networks from Decentralized Data,''
in \emph{Proc. 20th International Conference on Artificial Intelligence and Statistics (AISTATS)},
vol. 54, pp. 1273--1282, 2017.

\bibitem{blanchard2017krum}
P. Blanchard, E. M. El Mhamdi, R. Guerraoui, and J. Stainer,
``Machine Learning with Adversaries: Byzantine Tolerant Gradient Descent,''
in \emph{Proc. Advances in Neural Information Processing Systems (NeurIPS)},
vol. 30, pp. 119--129, 2017.

\bibitem{yin2018median}
D. Yin, Y. Chen, R. Kannan, and P. Bartlett,
``Byzantine-Robust Distributed Learning: Towards Optimal Statistical Rates,''
in \emph{Proc. 35th International Conference on Machine Learning (ICML)},
vol. 80, pp. 5650--5659, 2018.

\bibitem{elmhamdi2018bulyan}
E.-M. El-Mhamdi, R. Guerraoui, and S. Rouault,
``The Hidden Vulnerability of Distributed Learning in Byzantium,''
in \emph{Proc. 35th International Conference on Machine Learning (ICML)},
vol. 80, pp. 3521--3530, 2018.

\bibitem{cao2021fltrust}
X. Cao, M. Fang, J. Liu, and N. Z. Gong,
``FLTrust: Byzantine-Robust Federated Learning via Trust Bootstrapping,''
in \emph{Proc. Network and Distributed System Security Symposium (NDSS)},
2021.

\bibitem{baruch2019alie}
G. Baruch, M. Baruch, and Y. Goldberg,
``A Little Is Enough: Circumventing Defenses for Distributed Learning,''
in \emph{Proc. Advances in Neural Information Processing Systems (NeurIPS)},
vol. 32, pp. 8635--8645, 2019.

\bibitem{xie2020ipm}
C. Xie, O. Koyejo, and I. Gupta,
``Fall of Empires: Breaking Byzantine-Tolerant SGD by Inner Product Manipulation,''
in \emph{Proc. 35th Conference on Uncertainty in Artificial Intelligence (UAI)},
vol. 115, pp. 261--270, 2020.

\bibitem{otsuka2026demoa}
K. Otsuka, Y. Takezawa, and M. Yamada,
``Delayed Momentum Aggregation: Communication-efficient Byzantine-robust Federated Learning with Partial Participation,''
in \emph{Proc. International Conference on Machine Learning (ICML)},
2026.

\bibitem{lugosi2019meanestimation}
G. Lugosi and S. Mendelson,
``Mean Estimation and Regression Under Heavy-Tailed Distributions: A Survey,''
\emph{Foundations of Computational Mathematics},
vol. 19, no. 5, pp. 1145--1190, 2019.

\bibitem{bonneel2015swb}
N. Bonneel, J. Rabin, G. Peyr\'e, and H. Pfister,
``Sliced and Radon Wasserstein Barycenters of Measures,''
\emph{Journal of Mathematical Imaging and Vision},
vol. 51, no. 1, pp. 22--45, 2015.

\end{thebibliography}
\end{document}